\def\year{2020}\relax
\documentclass[letterpaper]{article} 
\usepackage{aaai20MOD}  
\usepackage{times}  
\usepackage{helvet} 
\usepackage{courier}  
\usepackage[hyphens]{url}  
\usepackage{graphicx} 
\usepackage{graphicx}  
\title{What Do You Mean I’m Funny? Personalizing the Joke Skill of a Voice-Controlled Virtual Assistant}

\author{\Large  \textbf{Alejandro Mottini and Amber Roy Chowdhury}\\ 
Amazon\\ 
Seattle\\
USA\\
(amottini,amberch)@amazon.com 
}

\begin{document}

\maketitle

\begin{abstract}
A considerable part of the success experienced by Voice-controlled virtual assistants (VVA) is due to the emotional and personalized experience they deliver, with humor being a key component in providing an engaging interaction. In this paper we describe methods used to improve the joke skill of a VVA through personalization. The first method, based on traditional NLP techniques, is robust and scalable. The others combine self-attentional network and multi-task learning to obtain better results, at the cost of added complexity. A significant challenge facing these systems is the lack of explicit user feedback needed to provide labels for the models. Instead, we explore the use of two implicit feedback-based labelling strategies. All models were evaluated on real production data. Online results show that models trained on any of the considered labels outperform a heuristic method, presenting a positive real-world impact on user satisfaction. Offline results suggest that the deep-learning approaches can improve the joke experience with respect to the other considered methods.
\end{abstract}

\section{Introduction}
\label{intro}

Voice-controlled virtual assistants (VVA) such as Siri and Alexa have experienced an exponential growth in terms of number of users and provided capabilities. They are used by millions for a variety of tasks including shopping, playing music, and even telling jokes. Arguably, their success is due in part to the emotional and personalized experience they provide. One important aspect of this emotional interaction is humor, a fundamental element of communication. Not only can it create in the user a sense of personality, but also be used as fallback technique for out-of-domain queries \cite{Bellegarda2014}. Usually, a VVA's humorous responses are invoked by users with the phrase \textit{"Tell me a joke"}. In order to improve the joke experience and overall user satisfaction with a VVA, we propose to personalize the response to each request. To achieve this, a method should be able to recognize and evaluate humor, a challenging task that has been the focus of extensive work. Some authors have applied traditional NLP techniques \cite{yan2017s}, while others deep learning models \cite{donahue2017humorhawk}. Moreover, \cite{yang2015humor} follows a semantic-based approach, while \cite{ruch1992assessment} and \cite{ahuja2018makes} tackle the challenge from a cognitive and linguistic perspective respectively.

To this end, we have developed two methods. The first one is based on traditional NLP techniques. Although relatively simple, it is robust, scalable, and has low latency, a fundamental property for real-time VVA systems.  The other approaches combine multi-task learning \cite{caruana1997multitask} and self-attentional networks \cite{vaswani2017attention} to obtain better results, at the cost of added complexity. Both BERT \cite{devlin2018bert} and an adapted transformer \cite{vaswani2017attention} architecture are considered. This choice of architecture was motivated by the advantages it presents over traditional RNN and CNN models, including better performance \cite{liu2018generating}, faster training/inference (important for real-time systems), and better sense disambiguation \cite{tang2018self} (an important component of computational humor \cite{yang2015humor}).

The proposed models use binary classifiers to perform point-wise ranking, and therefore require a labelled dataset. To generate it, we explore two implicit user-feedback labelling strategies: five-minute reuse and one-day return. Online A/B testing is used to determine if these labelling strategies are suited to optimize the desired user-satisfaction metrics, and offline data to evaluated and compared the system's performance.

\section{Method}
\label{model}


\subsection{Labelling Strategies}
\label{labeling}


Generating labels for this VVA skill is challenging. Label generation through explicit user feedback is unavailable since asking users for feedback creates friction and degrade the user experience. In addition, available humor datasets such as \cite{yang2015humor,potash2017semeval} only contain jokes and corresponding labels, but not the additional features we need to personalize the jokes.  

To overcome this difficulty, it is common to resort to implicit feedback. In particular, many VVA applications use interruptions as negative labels, the rationale being that unhappy users will stop the VVA. This strategy, however, is not suitable for our use-case since responses are short and users need to hear the entire joke to decide if it is funny. Instead, we explore two other implicit feedback labelling strategies: five-minute reuse and 1-day return.  Five-minute reuse labels an instance positive if it was followed by a new joke request within five-minutes. Conversely, 1-day return marks as positive all joke requests that were followed by a new one within the following 1 to 25-hour interval. Both strategies assume that if a user returns, he is happy with the jokes. This is clearly an approximation, since a returning user might be overall satisfied with the experience, but not with all the jokes. The same is true for the implied negatives; the user might have been satisfied with some or all of the jokes. Therefore, these labels are noisy and only provide weak supervision to the models.

Table \ref{exLabels} shows an example of the labels' values for a set of joke requests from one user.

\begin{table}[t!]
\caption{\label{exLabels} Example of labelling strategies: five-minute reuse (label 1) and 1-day return (label 2)}
  \begin{center}
  \begin{tabular}{|c|c|c|}
    \hline
     \bf Timestamp & \bf Label 1 & \bf Label 2 \\
   \hline
     2019/05/03-17:51:10& 1& 0\\
 2019/05/03-17:53:10& 0& 0\\
2019/05/06-21:41:09& 1& 1\\
 2019/05/06-21:44:19& 0& 1\\
2019/05/07-20:34:19& 0 & 0 \\
\hline
\end{tabular}
\end{center}

\end{table}

\subsection{Features}
\label{features}

All models have access to the same raw features, which we conceptually separate into user, item and contextual features. Examples of features in each of these categories are shown in Table \ref{featuresRaw}. Some of these are used directly by the models, while others need to be pre-processed. The manner in which each model consumes them is explained next.

\begin{table}
 \caption{\label{featuresRaw} Examples of features within each category}
   \begin{center}
  \begin{tabular}{|c|c|c|}
\hline
    \bf  Feature& \bf  Type & \bf  Category \\
\hline
    Country Code & Categorical & User\\
    Joke Text & String& Item \\
    Request Timestamp & Timestamp& Context\\
\hline
\end{tabular}
\end{center}

\end{table}


\subsection{NLP-based: LR-Model}
\label{prodSys}


To favor simplicity over accuracy, a logistic regression (LR) model is first proposed. Significant effort was put into finding expressive features. Categorical features are one-hot encoded and numerical ones are normalized. The raw Joke Text and Timestamp features require special treatment. The Joke Text is tokenized and the stop-words are removed. We can then compute computational humor features on the clean text such as sense combination \cite{yang2015humor} and ambiguity \cite{mihalcea2007characterizing}. In addition, since many jokes in our corpus are related to specific events (Christmas, etc), we check for keywords that relate the jokes to them. For example, if "Santa" is included, we infer it is a Christmas joke.  Finally, pre-computed word embeddings with sub-word information are used to represent jokes by taking the average and maximum vectors over the token representations. Sub-word information is important when encoding jokes since many can contain out-of-vocabulary tokens. The joke's vector representations are also used to compute a summarized view of the user's past liked and disliked jokes. We consider that a user liked a joke when the assigned label is 1, an approximation given the noisy nature of the labels. The user's liked/disliked joke vectors are also combined with the candidate joke vector by taking the cosine similarity between them.

For the raw Timestamp feature, we first extract simple time/date features such as month, day and isWeekend. We then compute binary features that mark if the timestamp occurred near one of the special events mentioned before. Some of these events occur the same day every year, while others change (for example, the Super Bowl). In addition, many events are country dependent.  The timestamp's event features are combined with the joke's event features to allow the model to capture if an event-related joke occurs at the right time of the year.

The LR classifier is trained on the processed features and one of the labels. The model's posterior probability is used to sort the candidates, which are chosen randomly from a pool of unheard jokes.  Although useful (see Validation section), this model has several shortcomings. In particular, many of the used features require significant feature engineering and/or are country/language dependent, limiting the extensibility of the model.

\subsection{Deep-Learning-based: DL-Models}
\label{DL}

\begin{figure}
\includegraphics[scale=0.5]{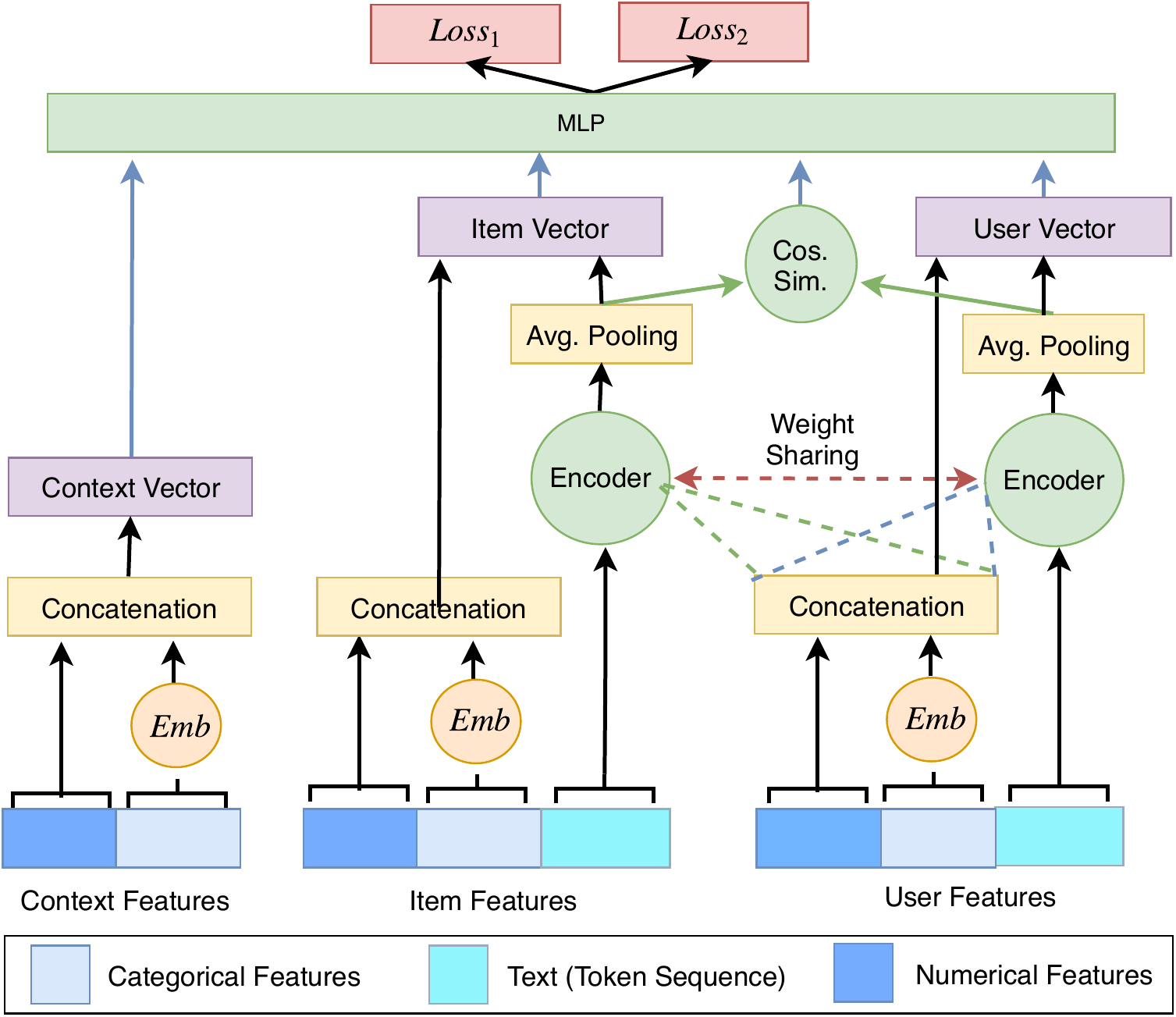}
\caption{Architecture of the transformer-based model}
\label{architectureGlobal}
\end{figure}

To overcome the LR-model's limitations, we propose the following model (see Figure \ref{architectureGlobal}). In the input layer, features are separated into context, item and user features. Unlike the LR-model, time and text features do not require extensive feature engineering. Instead, simple features (day, month and year) are extracted from the timestamp. After tokenization and stop-word removal, text features are passed through a pre-trained word embeding layer, and later, input into the joke encoder block.
The basis of the joke encoder is a modified transformer. Firstly, only the encoder is needed. Moreover, since studies suggest that humor is subjective and conditioned on the user's context \cite{8e165aa5efb347fc937dc3189dbb4fb3}, we add an additional sub-layer in the transformer encoder that performs attention over the user's features. This sub-layer, inserted between the two typical transformer sub-layers at certain depths of the network, allows the encoder to adapt the representations of the jokes to different user contexts. Thus, the same joke can be encoded differently depending on the user's features. In practice, this additional sub-layer works like the normal self-attention sub-layer, except it creates its query matrix Q from the sub-layer below, and its K and V matrices from the user features. As an alternative, we also test encoding the jokes using a pre-trained BERT model. 

Regardless of the used encoder, we average the token representations to obtain a global encoding of the jokes. The same encoder is used to represent the item's (the joke to rank) and the user's (liked and disliked jokes) textual features through weight sharing, and the cosine similarity between both representations are computed. The processed features are then concatenated and passed through a final block of fully connected layers that contains the output layers. Since experiments determined (see Validation section) that both labeling strategies can improve the desired business metrics, instead of optimizing for only one of them, we take a multi-task learning approach. Thus, we have two softmax outputs.


Finally, we use a loss function that considers label uncertainty, class imbalance and the different labeling functions. We start from the traditional cross-entropy loss for one labelling function. We then apply uniform label smoothing \cite{szegedy2016rethinking}, which converts the one-hot-encoded label vectors into smoothed label vectors towards $0.5$:
\begin{equation}
y_{ls}= y_{one-hot}*(1-\epsilon) + \frac{\epsilon}{2}
\end{equation}
with $\epsilon$ a hyper-parameter. Label smoothing provides a way of considering the uncertainty on the labels by encouraging the model to be less confident. We have also experimented with other alternatives, including specialized losses such as \cite{martinez2018taming}. However, they did not produce a significant increase in performance in our tests. To further model the possible uncertainty in the feedback, we apply sample weights calculated using an exponential decay function on the time difference between the current and the following training instance of the same customer:
\begin{equation}
w_i= a*b^{t_{i}}+1.0
\end{equation}
where $w_i$ is the weight of sample $i$, $t_i$ is the time difference between instances $i$ and $i+1$ for the same user, and $a,b$ are hyper-parameters such that $a>0$ and $0<b<1$. The rationale behind these weights is the following. If for example, we consider labeling function 1, and a user asks for consecutive jokes, first within 10 seconds and later within 4.9 minutes, both instances are labeled as positive. However, we hypothesize that there is a lower chance that in the second case the user requested an additional joke because he liked the first one. In addition, class weights are applied to each sample to account for the natural class imbalance of the dataset. Finally, the total loss to be optimized is the weighted sum of the losses for each of the considered labeling functions:
\begin{equation}
\mathcal{L}((f(x),\Theta),y) = \sum_{l=1}^2 w_l \mathcal{L}_{l}
\end{equation}
where $w_{l}$  are manually set weights for each label and $\mathcal{L}_{l}$ are the losses corresponding to each label, which include all the weights mentioned before.


\section{Validation}
\label{validation}

A two-step validation was conducted for English-speaking customers. An initial A/B testing for the LR model in a production setting was performed to compare the labelling strategies. A second offline comparison of the models was conducted on historical data and a selected labelling strategy. One month of data and a subset of the customers was used (approx. eighty thousand). The sampled dataset presents a fraction of positive labels of approximately 0.5 for reuse and 0.2 for one-day return. Importantly, since this evaluation is done on a subset of users, the dataset characteristic's do not necessarily represent real production traffic. The joke corpus in this dataset contains thousands of unique jokes of different categories (sci-fi, sports, etc) and types (puns, limerick, etc). The dataset was split timewise into training/validation/test sets, and hyperparameters were optimized to maximize the AUC-ROC on the validation set. As a benchmark, we also consider two additional methods: a non-personalized popularity model and one that follows \cite{kim2014convolutional}, replacing the transformer joke encoder with a CNN network (the specialized loss and other characteristics of the DL model are kept).

Hyperparameters were optimized using grid-search for the LR-Model. Due to computational constraints, random search was instead used for the DL-Model. In both cases, hyperparameters are selected to optimize the AUC-ROC on the validation set. Table \ref{paramsLR} lists some of the considered hyperparameter values and ranges for both models.  The actual optimal values are sample specific.

\begin{table}[t!]
 \caption{ \label{paramsLR} Hyperparameter values tuned over, LR (top) and DL models (bottom)}
  \begin{center}
  \begin{tabular}{|c|c|}
\hline
    Name & Value \\
\hline
Elastic-Net param. & [0.01,0.5] \\  
Regularization  &  [$10^{-3}$,10.0]  \\  
Fit intercept & True/False \\
\hline
Learning rate  &  [$10^{-3},10^{-5}$] \\  
Batch size & [32,256] \\
Label smoothing & [0.1,0.3] \\
Keep probability  & [0.5,0.8]  \\
Num. heads & [2,6] \\
Num. transformer layers & [1,6] \\
F.C layers& [2,5] \\
F.C layer sizes& [16,256] \\
CNN filter sizes & [2,32] \\
CNN num. filters & [16,128]\\
\hline
\end{tabular}
\end{center}

\end{table}

\subsection{Online Results: A/B Testing}
\label{ab}

Two treatment groups are considered, one per label. Users in the control group are presented jokes at random, without repetition. Several user-satisfaction metrics such as user interruption rate, reuse of this and other VVA skills, and number of active dialogs are monitored during the tests. The relative improvement/decline of these metrics is compared between the treatments and control, and between the treatments themselves. The statistical significance is measured when determining differences between the groups. Results show that the LR-based model consistently outperforms the heuristic method for both labeling strategies, significantly improving retention, dialogs and interruptions. These results suggest that models trained using either label can improve the VVA's joke experience.

\subsection{Offline Results}
\label{results}

One-day return was selected for the offline evaluation because models trained on it have a better AUC-ROC, and both labeling strategies were successful in the online validation. All results are expressed as relative change with respect to the popularity model.

\begin{table}[t!]
\caption{\label{res1} Relative change w.r.t popularity model of AUC-ROC and Overall Accuracy: transformer model (DL-T), BERT model (DL-BERT), transformer without special context-aware attention (DL-T-noAtt) and without both special attention and modified loss (DL-T-basic), CNN model (DL-CNN) and LR model (LR).}
  \begin{center}
  \begin{tabular}{|c|c|c|}
\hline
    Method & R. Ch. AUC-ROC & R. Ch.  O.A. \\
\hline
DL-T           & 0.31 &  0.24   \\   
DL-BERT           & 0.30 &  0.27   \\   
DL-T-noAtt & 0.29 &   0.24  \\ 
DL-T-basic & 0.28 &  0.23   \\    
DL-CNN         & 0.28 &  0.13   \\  
LR             & 0.24 &  0.21  \\  
\hline
\end{tabular}
\end{center}

\end{table}
We start by evaluating the models using AUC-ROC. As seen in Table \ref{res1}, the transformer-based models, and in particular our custom architecture, outperform all other approaches. Similar conclusions can be reached regarding overall accuracy. However, given the class imbalance, accuracy is not necessarily the best metric to consider. In addition, to better understand the effect to the original transformer architecture, we present the performance of the model with and without the modified loss and special attention sub-layer (see Table \ref{res1}). Results suggest both modifications have a positive impact on the performance. Finally, to further evaluate the ranking capabilities of the proposed methods, we use top-1 accuracy. Additional positions in the ranking are not considered because only the top ranked joke is presented to the customer. Results show that the DL based models outperform the other systems, with a relative change in top-1 accuracy of 1.4 for DL-BERT and 0.43 for DL-T, compared with 0.14 for the LR method.

Results show that the proposed methods provide different compromises between accuracy, scalability and robustness. On one hand, the relatively good performance of the LR model with engineered features provides a strong baseline both in terms of accuracy and training/inference performance, at the cost of being difficult to extend to new countries and languages. On the other hand, DL based methods give a significant accuracy gain and require no feature engineering, which facilitates the expansion of the joke experience to new markets and languages. This comes at a cost of added complexity if deployed in production. In addition, given the size of the BERT model (340M parameters), real-time inference using DL-BERT becomes problematic due to latency constraints. In this regard, the DL-T model could be a good compromise since its complexity can be adapted, and it provides good overall accuracy.

\section{Conclusions and Future Work}
\label{conclusions}

This paper describes systems to personalize a VVA's joke experience using NLP and deep-learning techniques that provide different compromises between accuracy, scalability and robustness. Implicit feedback signals are used to generate weak labels and provide supervision to the ranking models. Results on production data show that models trained on any of the considered labels present a positive real-world impact on user satisfaction, and that the deep learning approaches can potentially improve the joke skill with respect to the other considered methods. In the future, we would like to compare all methods in A/B testing, and to extend the models to other languages.




\bibliography{bioV1}
\bibliographystyle{aaai}


\end{document}